\documentclass[11pt]{article}

\usepackage{cmap}
\usepackage[utf8]{inputenc}
\usepackage[T1]{fontenc}
\usepackage{lmodern}
\usepackage[protrusion=true,expansion=false]{microtype}
\usepackage[margin=1in, top=1.1in, bottom=1.1in]{geometry}

\usepackage{amsmath,amssymb,amsthm}

\usepackage{booktabs}
\usepackage{makecell}
\usepackage{graphicx}
\usepackage{caption}
\usepackage[colorlinks=true,
            linkcolor=blue!70!black,
            citecolor=blue!70!black,
            urlcolor=blue!70!black]{hyperref}

\usepackage{xcolor}
\usepackage{enumitem}
\usepackage[numbers,sort&compress]{natbib}
\usepackage{tikz}
\usetikzlibrary{arrows.meta,positioning,fit,calc,backgrounds,shapes.geometric,
                 decorations.pathreplacing,patterns}
\usepackage{pgfplots}
\pgfplotsset{compat=1.18}
\usepackage{array}
\usepackage{multirow}
\usepackage{float}

\usepackage{titlesec}
\titleformat{\section}{\large\bfseries}{\thesection}{1em}{}
\titleformat{\subsection}{\normalsize\bfseries}{\thesubsection}{1em}{}

\newtheorem{theorem}{Theorem}[section]

\newtheorem{corollary}[theorem]{Corollary}
\newtheorem{proposition}[theorem]{Proposition}
\newtheorem{definition}[theorem]{Definition}

\newtheorem{observation}[theorem]{Observation}

\newcommand{\allow}{\texttt{Allow}}
\newcommand{\deny}{\texttt{Deny}}
\newcommand{\escalate}{\texttt{Escalate}}
\newcommand{\Agents}{\mathcal{A}}
\newcommand{\Actors}{\mathcal{U}}
\newcommand{\Azero}{\mathcal{A}_0}
\newcommand{\Azeroi}[1]{\mathcal{A}_{0,#1}}   
\newcommand{\Dhat}{\widehat{D}}
\newcommand{\State}[1]{S_{#1}}
\newcommand{\Trans}[1]{\delta_{#1}}
\newcommand{\Sum}[1]{\lambda_{#1}}
\newcommand{\Layer}[1]{L_{#1}}
\newcommand{\Comp}{\otimes_\kappa}
\newcommand{\AsyncComp}{\xrightarrow{\kappa}}
\newcommand{\join}{\sqcup}
\newcommand{\Share}[1]{S_{#1}}
\newcommand{\Dec}{\mathcal{D}}
\newcommand{\kbound}{k_0}
\newcommand{\Kcap}{K}

\begin{document}

\title{\textbf{Irreducible Governance Structure for Autonomous Agent Systems:\\
Fair Allocation, Strategy-Proofness, and Multi-Scale Composition}}

\author{Marcelo Fernandez\\
TraslaIA\\
\texttt{info@traslaia.com}}

\date{2026}

\maketitle

\begin{table}[h]
\centering
\small
\caption{The Agent Governance Series. This paper consolidates Papers P3 and P4.}
\label{tab:series}
\begin{tabular}{@{}llll@{}}
\toprule
\textbf{Paper} & \textbf{Title} & \textbf{Zenodo DOI} & \textbf{arXiv} \\
\midrule
P0   & Atomic Decision Boundaries~\cite{fernandez2026a}         & 10.5281/zenodo.19670649 & 2604.17511 \\
P1   & Agent Control Protocol (ACP)~\cite{fernandez2026b}       & 10.5281/zenodo.19672575 & 2603.18829 \\
P2   & From Admission to Invariants (IML)~\cite{fernandez2026c} & 10.5281/zenodo.19672589 & 2604.17517 \\
\textbf{P3/4} & \textbf{Irreducible Governance Structure} (this paper) & 10.5281/zenodo.19708496 & TBD \\
P5   & Reconstructive Authority Model (RAM)~\cite{fernandez2026ram} & 10.5281/zenodo.19669430 & 2604.22898 \\
P6   & Operationalizing Reconstructive Authority~\cite{fernandez2026op} & 10.5281/zenodo.19699460 & 2605.23935 \\
P7   & Closing the Execution Gap & 10.5281/zenodo.19929771 & --- \\
P8   & Identity-Bound Governance & 10.5281/zenodo.20157139 & 2610.00787 \\
\bottomrule
\end{tabular}
\end{table}

\begin{abstract}
Autonomous agent governance is commonly framed as a problem of enforcing correct execution. However, in multi-agent systems, an equally fundamental question precedes execution: which agents are allowed to act.

In this paper, we show that allocation is not an auxiliary concern but a structural component of governance. When multiple agents compete for execution under constrained resources, any allocation mechanism introduces unavoidable trade-offs between fairness, efficiency, and resistance to strategic manipulation. We formalize this through a set of impossibility results, including Sybil amplification and the incompatibility of desirable allocation properties.

We then show that these allocation constraints cannot be resolved within a single layer of control. Instead, they propagate across the system, interacting with execution atomicity, state enforcement, and behavioral drift. This leads to a multi-scale governance structure composed of four orthogonal dimensions: temporal (atomicity), state (enforcement), behavioral (drift monitoring), and population (allocation).

Our main result is an irreducibility theorem: under finite observability, no governance architecture can collapse these dimensions without loss of correctness or stability. Allocation, in particular, acts as a first-class constraint that prevents simplification of the governance stack.

We validate these results through a series of analytical constructions and empirical ablations, showing that removing or approximating any layer leads to systematic failure modes.

This work reframes agent governance as a structural problem: not only how actions are executed, but how access to execution is determined, and why these decisions impose irreducible architectural requirements.
\end{abstract}

\tableofcontents
\newpage

\section{Introduction}
\label{sec:intro}

Autonomous agent systems are typically analyzed through the lens of execution: how to ensure that actions are correct, consistent, and compliant with specified policies. Prior work has established that this problem is non-trivial even in isolation. Ensuring atomicity under dynamic conditions is structurally impossible when evaluation and execution are separated~\cite{fernandez2026a}. Enforcement mechanisms can guarantee state-level constraints but remain blind to behavioral drift~\cite{fernandez2026c}. As a result, governance requires multiple interacting layers to maintain consistency under limited observability.

However, these analyses implicitly assume that the system has already decided which actions will be executed.

In multi-agent settings, this assumption does not hold.

When multiple agents compete for execution under shared constraints---whether computational resources, access privileges, or external impact budgets---the system must decide which agents are allowed to reach the execution boundary. This introduces an allocation problem that precedes and conditions all subsequent governance layers.

A common approach is to treat allocation as a secondary concern, handled through scheduling, prioritization, or heuristic policies. In this paper, we argue that this view is incomplete. Allocation is not an implementation detail; it is a structural component of governance.

We show that any allocation mechanism in such systems is subject to fundamental constraints. In particular, when agents can adapt their behavior or identity, the system becomes vulnerable to amplification effects (e.g., Sybil strategies), and desirable properties such as fairness, efficiency, and strategy-proofness cannot be simultaneously satisfied. These are not artifacts of specific designs but arise from the underlying structure of the problem.

Crucially, these constraints do not remain confined to the allocation layer. They propagate across the system. Decisions about which agents are admitted to execution affect the validity of atomic guarantees, interact with enforcement mechanisms, and shape the conditions under which behavioral drift manifests.

This leads to the central claim of this paper: agent governance is inherently multi-scale, and its structure is irreducible.

We formalize this by defining four orthogonal dimensions of governance:
\begin{itemize}
  \item \textbf{Temporal} --- ensuring atomic execution under changing conditions (Paper~0~\cite{fernandez2026a})
  \item \textbf{State} --- enforcing constraints over system state (Paper~1~\cite{fernandez2026b})
  \item \textbf{Behavioral} --- detecting and managing drift over time (Paper~2~\cite{fernandez2026c})
  \item \textbf{Population} --- allocating access to execution among competing agents (this paper)
\end{itemize}

Each dimension addresses a distinct failure mode. None can be eliminated or subsumed by another without introducing new vulnerabilities.

Our main result is an irreducibility theorem: under finite observability, no governance architecture can collapse these dimensions into a simpler form while preserving correctness and stability. In particular, the population dimension introduces constraints that cannot be absorbed into execution or enforcement layers.

We support this result through formal analysis and empirical validation. Using both theoretical constructions and ablation studies, we show that removing or approximating any dimension leads to systematic failure modes.

The contribution of this work is twofold. First, we establish allocation as a first-class governance dimension with its own formal constraints. Second, we show that these constraints imply an irreducible multi-layer architecture for autonomous agent systems.

This reframes the problem of governance: it is not only about ensuring that actions are correct, but about structuring the system so that only admissible actions can occur, and understanding why this structure cannot be simplified.

\section{System Model: Multi-Agent Governance}
\label{sec:model}

\subsection{Multi-Agent Labeled Transition System}

We build on the execution model of \cite{fernandez2026a}. Let $M = (S, \mathit{Act}, {\to})$ be a labeled transition system (LTS), where $S$ is a state space, $\mathit{Act}$ is an action alphabet partitioned into agent actions $A_{\mathrm{ag}}$ and environment actions $A_{\mathrm{env}}$.

\begin{definition}[Agent Population]
\label{def:agents}
An \emph{agent population} is a finite set $\Agents = \{a_1, \ldots, a_N\}$. Each agent $a_i \in \Agents$ generates action requests from a per-agent action alphabet $\mathrm{Act}_i \subseteq A_{\mathrm{ag}}$.
\end{definition}

\begin{definition}[Actor Mapping]
\label{def:actors}
An \emph{actor set} is a finite set $\Actors = \{u_1, \ldots, u_M\}$ with $M \leq N$. The \emph{actor mapping} $\pi : \Agents \to \Actors$ assigns each agent to its controlling actor. For actor $u_j \in \Actors$, we write $\Agents_j = \pi^{-1}(u_j)$ for the set of agents it controls, with $|\Agents_j| = m_j$.
\end{definition}

\subsection{Atomic Decision Boundary and Allocation}

The atomic decision function $F : S \times A_{\mathrm{ag}} \to \Dec \times S$ (from Paper~0) is unchanged. In the multi-agent setting, multiple agents may present requests to $F$ concurrently. Since $F$ is a single indivisible LTS step, concurrent requests must be serialized.

\begin{definition}[Serialized Atomic Boundary with Allocation]
\label{def:serial-alloc}
A \emph{serialized atomic decision boundary} for population $\Agents$ is a pair $(F, \alpha)$ where:
\begin{enumerate}[label=(\roman*)]
  \item $F : S \times A_{\mathrm{ag}} \to \Dec \times S$ is an atomic decision function;
  \item $\alpha : \mathrm{PendingRequests}(\Agents) \times S \to \Agents$ is an \emph{allocation function} that selects the next agent whose request is presented to $F$.
\end{enumerate}
The decision space $\Dec = \{\allow, \escalate, \deny\}$ is equipped with a partial order $\allow \prec \escalate \prec \deny$. The join $\join$ denotes the least upper bound.
\end{definition}

\subsection{Per-Agent Bounded Enforcement}

Each agent operates under a bound $\kbound \in \mathbb{N}$ on the number of admissible actions it may execute within an enforcement context.

\begin{definition}[Global Capacity]
\label{def:globalcap}
For a population $\Agents$ of $N$ agents each with bound $\kbound$, the \emph{global admissible capacity} is:
\[
  \Kcap \;=\; N \cdot \kbound.
\]
\end{definition}

\begin{definition}[Execution Share and Actor Share]
\label{def:shares}
For a finite execution trace, the \emph{execution share} of agent $a_i$ is:
\[
  \Share{i} \;=\; \frac{\kappa_i}{\sum_{j=1}^{N} \kappa_j},
\]
where $\kappa_i$ is the count of $\allow$ decisions issued to $a_i$. The \emph{actor share} is:
\[
  \Share{u_j} \;=\; \sum_{a_i \in \Agents_j} \Share{i}.
\]
\end{definition}

\section{Allocation Layer: Fairness and Structural Constraints}
\label{sec:allocation}

\subsection{Fairness Definitions}

We introduce a hierarchy of fairness notions for atomic governance, ordered by strength.

\begin{definition}[Share Fairness]
\label{def:share-fair}
A system satisfies \emph{$\varepsilon$-share fairness} if for all agents $a_i \in \Agents$:
\[
  \Bigl|\,\Share{i} - \tfrac{1}{N}\,\Bigr| \;\leq\; \varepsilon.
\]
\end{definition}

\begin{definition}[Actor-Level Proportionality]
\label{def:actor-fair}
A system satisfies \emph{$\varepsilon$-actor-level proportionality} if for all actors $u_j \in \Actors$:
\[
  \Bigl|\,\Share{u_j} - \tfrac{1}{M}\,\Bigr| \;\leq\; \varepsilon.
\]
\end{definition}

\begin{definition}[Envy-Freeness]
\label{def:envy}
An allocation is \emph{envy-free} if no agent strictly prefers another agent's allocation under identical valuations.
\end{definition}

\begin{definition}[Strategy-Proofness Against Fragmentation]
\label{def:strategy-proof}
An allocation mechanism $\alpha$ is \emph{strategy-proof against identity fragmentation} if, for any actor $u_j$ controlling $m_j$ agents:
\[
  \frac{\partial\, \Share{u_j}}{\partial\, m_j} \;=\; 0.
\]
Splitting an agent into multiple agents under the same actor does not increase the actor's share.
\end{definition}

\begin{definition}[Starvation Freedom]
\label{def:starvation}
A system is \emph{starvation-free} if every agent with a pending admissible request eventually receives an $\allow$ decision.
\end{definition}

\subsection{Structural Failure Modes}

\begin{definition}[Sybil Amplification]
\label{def:sybil}
A system exhibits \emph{Sybil amplification} if actor share $\Share{u_j}$ increases with $m_j = |\Agents_j|$:
\[
  \exists\, u_j \in \Actors : \quad
  \Share{u_j}(m_j + 1) \;>\; \Share{u_j}(m_j).
\]
\end{definition}

\begin{definition}[Temporal Domination]
\label{def:temporal-dom}
A system exhibits \emph{temporal domination} if there exist agents $a_i, a_j$ and times $t_1 < t_2$ such that: $a_i$ arrives at $t_1$, $a_j$ arrives at $t_2$, yet $a_i$ exhausts capacity while $a_j$ receives nothing despite having admissible pending requests.
\end{definition}

\begin{definition}[Resource Contention Unfairness]
\label{def:contention}
A system exhibits \emph{resource contention unfairness} if, in the absence of an explicit allocation policy, agents' shares for shared resources exhibit persistent skew due to non-identity factors (timing, retry rate, latency).
\end{definition}

\subsection{Main Results: Impossibility Theorems}

\begin{theorem}[Sybil Amplification]
\label{thm:sybil}
Let $\Agents$ be a population of $N$ agents with per-agent bound $\kbound$, global capacity $\Kcap = N \cdot \kbound$, and actor mapping $\pi : \Agents \to \Actors$. Let the allocation function $\alpha$ be identity-oblivious (i.e., $\alpha$ does not use $\pi$). Then for any actor $u_j$ controlling $m_j$ agents:
\begin{enumerate}[label=(\roman*)]
  \item The actor share satisfies $\Share{u_j} = m_j / N$ under exact agent-level share fairness.
  \item There exists an assignment of $m_j$ and $N$ such that $\Share{u_j}$ is arbitrarily close to~$1$.
  \item No identity-oblivious allocation mechanism can guarantee $\varepsilon$-actor-level proportionality for $\varepsilon < \max_j (m_j/N - 1/M)$.
\end{enumerate}
\end{theorem}

\begin{proof}
(i) Under exact agent-level share fairness, each agent executes exactly $\kbound$ admitted actions. Therefore: $\Share{u_j} = \sum_{a_i \in \Agents_j} \frac{\kbound}{N \cdot \kbound} = \frac{m_j}{N}$.

(ii) Set $m_j = N - 1$. Then $\Share{u_j} = \frac{N-1}{N} \to 1$ as $N \to \infty$.

(iii) Since $\alpha$ is identity-oblivious, it cannot distinguish agents by actor and therefore cannot compensate for skew in $m_j$. An identity-oblivious mechanism that achieves $\varepsilon$-actor-level proportionality must have $\varepsilon \geq \max_j |m_j/N - 1/M|$.
\end{proof}

\begin{theorem}[Allocation Necessity]
\label{thm:necessity}
There exists a multi-agent system $(\Agents, F, \alpha_{\emptyset})$ in which $F$ satisfies the atomic decision boundary property and $\alpha_{\emptyset}$ is FCFS allocation, such that:
\begin{enumerate}[label=(\roman*)]
  \item Every admission decision by $F$ is atomically correct.
  \item The system violates actor-level proportionality with arbitrarily large deviation.
\end{enumerate}
Consequently, atomic correctness of $F$ does not imply any fairness property. An explicit allocation mechanism is a \emph{necessary additional component} of a fair governance system.
\end{theorem}

\begin{proof}
Construct $\Agents = \{a_1, a_2\}$ with all actions admissible everywhere. Under FCFS with $a_1$ arriving before $a_2$, let $a_1$ submit $T$ requests at time $t=0$ and $a_2$ submit $T$ requests at time $t=1$. With $\kbound = T$, we have $\kappa_1 = T$ and $\kappa_2 = 0$, violating actor-level proportionality ($|\Share{u_1} - 1/2| = 1/2$), while every individual decision is correct.
\end{proof}

\begin{theorem}[Strategy-Proofness Impossibility]
\label{thm:strategy}
Let $\alpha$ be an identity-oblivious allocation mechanism. Then $\alpha$ cannot simultaneously satisfy:
\begin{enumerate}[label=(\roman*)]
  \item $\varepsilon$-actor-level proportionality for any $\varepsilon < \max_j |m_j/N - 1/M|$; and
  \item strategy-proofness against identity fragmentation (Definition~\ref{def:strategy-proof}).
\end{enumerate}
Any actor-aware mechanism that achieves both properties must use $\pi$ to aggregate state across agents of the same actor, thereby breaking per-agent independence.
\end{theorem}

\begin{proof}
By Theorem~\ref{thm:sybil}(iii), identity-oblivious mechanisms cannot guarantee actor-level proportionality. Actor-aware mechanisms can achieve proportionality by tracking per-actor aggregate state, but this aggregation breaks per-agent independence: individual counter $\kappa_i$ must reference the aggregate state of $\Agents_j$, not just $a_i$'s local history.
\end{proof}

\subsection{Allocation Mechanisms}

\begin{definition}[Mechanisms M1--M4]
\label{def:mechanisms}
\begin{itemize}[noitemsep]
  \item \textbf{M1 (Per-Agent Token Bucket)}: FCFS among agents with $\kappa_i < \kbound$.
  \item \textbf{M2 (Round-Robin Fair Queuing)}: Cyclic order over agents, skipping exhausted quotas.
  \item \textbf{M3 (Actor-Aware Rate Limiting)}: Per-actor quota $K_U = \Kcap/M$; track aggregate $\kappa_{u_j} = \sum_{a_i \in \Agents_j} \kappa_i$.
  \item \textbf{M4 (Weighted Fair Queuing)}: Assign weights $w_i > 0$ to agents; serve in proportion to weights.
\end{itemize}
\end{definition}

\begin{proposition}[M2 achieves share fairness and starvation freedom]
\label{prop:m2}
Under M2, over any execution trace of length $T$ with all agents continuously submitting requests, the share deviation satisfies $|\Share{i} - 1/N| \leq 1/T$ for all $a_i$, and every agent with a pending admissible request eventually receives an $\allow$ decision.
\end{proposition}

\begin{proposition}[M3 achieves actor-level proportionality and strategy-proofness]
\label{prop:m3}
Under M3, for all actors $u_j \in \Actors$, the actor share satisfies $\Share{u_j} \leq 1/M$ regardless of $m_j$. Registering additional agents does not increase $\Share{u_j}$.
\end{proposition}

\begin{proposition}[M4 achieves actor-level proportionality asymptotically]
\label{prop:m4}
Under M4 with weights $w_i = 1/m_j$ for all $a_i \in \Agents_j$, the actor share converges: $\lim_{T \to \infty} \Share{u_j}(T) = 1/M$.
\end{proposition}

\section{From Allocation Constraints to Irreducible Multi-Scale Structure}
\label{sec:bridge}

The allocation constraints proved in Section~\ref{sec:allocation} do not remain confined to the allocation layer. They propagate upward and interact with every other layer of the governance architecture. This section establishes the crucial bridge: allocation is not merely a scheduling mechanism but a \emph{structural necessity} that imposes requirements on atomicity, enforcement, and monitoring.

\subsection{Allocation Constraints Propagate}

Consider a system equipped with atomic admission control (Paper~0) and stateful enforcement (Paper~1, ACP), but no allocation mechanism. By Theorem~\ref{thm:necessity}, such a system can be atomically correct at every decision point while violating fair allocation. The unfairness manifests not as faulty decisions but as systematic distributional skew.

Now add behavioral monitoring (Paper~2, IML). The IML framework detects drift via the Non-Identifiability Theorem: $\Azero \notin \sigma(g)$, where $\Azero$ is the admissible behavior contract and $g$ is the enforcement projection. Crucially, IML cannot distinguish between two failure modes:
\begin{enumerate}[noitemsep]
  \item \emph{Intrinsic drift}: the agent's true behavior has drifted from its admission-time contract.
  \item \emph{Allocation-induced drift}: the agent receives disproportionate access to the boundary (over-service) and explores state space regions it would not normally visit.
\end{enumerate}

Under over-service (e.g., one actor controlling $m$ agents under M1 or M2), an over-served agent's empirical tool distribution diverges from the admission-time expectation, triggering an IML drift signal $\Dhat_t > 0$ even though the agent's intrinsic behavior has not changed. This is not a bug in IML; it is a structural consequence of ignoring the allocation layer.

\begin{theorem}[Allocation Bias Induces IML Drift]
\label{thm:bias-drift}
Let $a_i \in \Agents$ be an agent calibrated under fair allocation ($B_i = 0$) with
admission-time snapshot $\Azeroi{i}$. Let $P_{\mathrm{excess}}$ be the distribution of
actions drawn from states visited only due to excess allocation, with
$\mathrm{JS}(P_{\mathrm{excess}} \| P_{\Azeroi{i}}) \geq \eta > 0$. If $a_i$ is
persistently over-served with bias $B_i(t) \geq \delta > 0$ over a window of $T$ steps,
then the IML deviation estimator satisfies
\[
  \widehat{D}(a_i,\,\Azeroi{i}) \;\geq\; 0.40 \cdot f^2 \eta / 2 \;>\; 0,
\]
where $f = \delta/(1/N + \delta)$ is the fraction of actions attributable to excess
allocation, even though $g(\tau_i) = 0$ (no individual action violates a local constraint).
\end{theorem}

\begin{proof}
By the IML deviation estimator $\Dhat_t = 0.40\,D_t + 0.35\,D_c + 0.25\,D_l$
(Paper~2~\cite{fernandez2026c}), the tool-distribution component $D_t$ dominates.
Under fraction $f$ of actions from $P_{\mathrm{excess}}$, the mixed empirical distribution
is at Jensen--Shannon distance $\geq f^2 \eta / 2$ from $P_{\Azeroi{i}}$. Since $f > 0$
and $\eta > 0$ by assumption, $D_t > 0$ and hence $\Dhat > 0$. The enforcement
projection $g(\tau_i)$ evaluates each action locally and cannot detect distributional
skew, so $g(\tau_i) = 0$ throughout. \hfill $\blacksquare$
\end{proof}

Conversely, under-service (e.g., an agent's allocation is artificially restricted) inflates
detection delay: fewer samples increase estimation error, making genuine drift harder to
detect. By Corollary of Theorem~\ref{thm:bias-drift}, detection delay is amplified by
$(1 - \delta N)^{-1}$ under under-service, and approaches infinity as starvation becomes
complete.

\subsection{The Population Dimension}

The allocation layer introduces a new dimension of observability: the \emph{population dimension}. Whereas the atomic layer observes single requests, the enforcement layer observes per-agent histories, and the behavioral layer observes traces, the allocation layer observes the \emph{distribution of access across agents}.

This distribution is not reducible to lower-layer observations. No function of per-agent histories can tell us whether one actor is receiving a disproportionate share---that determination requires aggregation across agent boundaries, which lower layers are explicitly designed not to do.

Moreover, identity fragmentation (Sybil attacks) is a population-level threat that cannot be mitigated by lower layers. Paper~0's atomic boundary has no mechanism to prevent splitting identities. Paper~1's ACP enforces per-agent bounds but is powerless against actor fragmentation. Paper~2's IML can detect distributional drift but cannot distinguish legitimate population heterogeneity from malicious Sybil amplification without access to the actor mapping $\pi$.

The allocation layer is therefore the \emph{minimal sufficient} layer to address population-level concerns. It operates at a scale (epoch-level, aggregating across agents) that is orthogonal to the request-level scale of lower layers.

\section{Multi-Scale Governance Architecture}
\label{sec:arch}

\subsection{Layer Definition}

We formalize governance as a composition of four transducers operating at distinct temporal and observability scales.

\begin{definition}[Layer]
\label{def:layer}
A \emph{governance layer} is a tuple
\[
  \Layer{i} = (\State{i},\, \Sigma_i,\, \Trans{i},\, \Sum{i}),
\]
where $\State{i}$ is a finite internal state space, $\Sigma_i$ is an input alphabet, $\Trans{i}: \State{i} \times \Sigma_i \to \State{i} \times M_i \times \Dec$ is the transition function, and $\Sum{i}: \State{i} \to \Lambda_i$ is the \emph{observable summary} function projecting internal state to a bounded summary space $\Lambda_i$.
\end{definition}

\subsection{The Four Layers}

\begin{itemize}[noitemsep]
  \item $\Layer{0}$: \textbf{Atomic boundary} (Paper~0). Operates \emph{per request}. Enforces that decision and state transition occur as a single LTS step.
  \item $\Layer{1}$: \textbf{ACP enforcement} (Paper~1). Operates \emph{per request}. Maintains per-agent state (Capability Token, Audit Ledger) under assumptions TA1 (serializable ledger) and TA2 (atomic commit).
  \item $\Layer{2}$: \textbf{IML behavioral monitor} (Paper~2). Operates \emph{per trace window} $W_t$ of $n$ consecutive requests. Produces deviation estimate $\Dhat_t = 0.40 D_t + 0.35 D_c + 0.25 D_l \in [0,1]$.
  \item $\Layer{3}$: \textbf{Fair allocator} (new contribution). Operates \emph{per epoch} $E_t$ (multiple windows). Receives $\Dhat_t$ and produces allocation $\alpha_t \in \Delta(\Agents)$.
\end{itemize}

\subsection{Synchronous and Asynchronous Composition}

$\Layer{0}$ and $\Layer{1}$ must act on every request, so they couple synchronously.

\begin{definition}[Synchronous Composition]
\label{def:sync}
The \emph{synchronous composition} $\Layer{0} \Comp \Layer{1}$ is the tuple $(\State{0} \times \State{1},\; \Sigma,\; \Trans{\otimes},\; \Sum{\otimes})$, where:
\begin{align*}
  (s_0', m_0, d_0) &= \Trans{0}(s_0, x), \\
  x_1 &= \kappa(m_0), \\
  (s_1', m_1, d_1) &= \Trans{1}(s_1, x_1), \\
  d_\otimes &= d_0 \join d_1,
\end{align*}
for coupling function $\kappa: M_0 \to \Sigma_1$ and joint summary $\Sum{\otimes}(s_0, s_1) = (\Sum{0}(s_0), \Sum{1}(s_1))$.
\end{definition}

$\Layer{1}$, $\Layer{2}$, and $\Layer{3}$ operate at different temporal scales (request, window, epoch), so they couple asynchronously.

\begin{definition}[Asynchronous Coupling]
\label{def:async}
The \emph{asynchronous coupling chain} $\Layer{1} \AsyncComp^{W_t}_\kappa \Layer{2} \AsyncComp^{E_t}_\kappa \Layer{3}$ is defined by:
\begin{enumerate}[label=(\roman*),noitemsep]
  \item $\Layer{2}$ receives the read-only window summary $\Sum{1}(W_t) \in \Lambda_1^n$ and produces $\Dhat_t \in [0,1]$.
  \item $\Layer{3}$ receives $\Dhat_t$ and produces $\alpha_t \in \Delta(\Agents)$.
\end{enumerate}
States remain isolated: $\State{i} \cap \State{j} = \emptyset$ for $i \neq j$, except $\Layer{2}$ has read-only access to $\Sum{1}(\State{1})$.
\end{definition}

\begin{definition}[Coupling Conditions]
\label{def:coupling}
The composed architecture satisfies:
\begin{enumerate}[label=(C\arabic*),noitemsep]
  \item \textit{Summary-only coupling}: $\kappa$ operates on $\Sum{i}$, not on full state $\State{i}$.
  \item \textit{State isolation with read-only access}: $\State{i} \cap \State{j} = \emptyset$ for $i \neq j$; $\Layer{2}$ reads $\Sum{1}(\State{1})$ but does not write to it.
  \item \textit{Monotonic decision aggregation}: $\join$ is monotone in severity.
\end{enumerate}
\end{definition}

\section{Irreducibility of the Multi-Scale Architecture}
\label{sec:irreducibility}

\subsection{Finite Observability Conditions}

\begin{definition}[Finite Observability]
\label{def:finite-obs}
An architecture is subject to \emph{finite observability} if:
\begin{enumerate}[label=(FO\arabic*),noitemsep]
  \item $|\State{i}| < \infty$ for all $i$ (finite state spaces),
  \item inter-layer communication is restricted to bounded summaries $\Sum{i}: \State{i} \to \Lambda_i$ where $|\Lambda_i| < \infty$,
  \item decisions are based on local observations and projected summaries only.
\end{enumerate}
\end{definition}

\subsection{The Four Guarantees}

\begin{definition}[Properties P1--P4]
\label{def:properties}
\begin{enumerate}[label=(P\arabic*),noitemsep]
  \item \textbf{P1 (Strong atomicity)}: Admission decisions enforce consistency at each request via atomic decision function $F$.
  \item \textbf{P2 (Behavioral non-identifiability bound)}: Enforcement projection $g(\tau)$ cannot distinguish all global behavioral contracts; IML monitor provides a bound via $\Dhat_t$.
  \item \textbf{P3 (Exact actor-level fairness)}: Allocation mechanism achieves per-agent fairness across heterogeneous population via actor-aware scheduling (M3/M4).
  \item \textbf{P4 (Sybil robustness)}: Adversarial identity fragmentation cannot amplify resource access.
\end{enumerate}
\end{definition}

\subsection{Main Theorems: Interface Compatibility and Irreducibility}

\begin{theorem}[Interface Compatibility]
\label{thm:compatibility}
Let $\Layer{0}$, $\Layer{1}$ be synchronously composed via $\Comp$, and let $\Layer{2}$, $\Layer{3}$ be asynchronously coupled with conditions C1--C3.

If (a) each layer $\Layer{i}$ preserves its local guarantee under its native operating scale, and (b) coupling mappings $\kappa$, $\Sum{i}$ do not collapse distinctions required by those guarantees, then the composed system preserves the guarantees of all layers at their respective operating scales.
\end{theorem}

\begin{proof}
\textbf{Synchronous composition preserves $\Layer{0}$ and $\Layer{1}$ guarantees.}
By Definition~\ref{def:sync}, the dynamics of $\Layer{0}$ in composition are $(s_0', m_0, d_0) = \Trans{0}(s_0, x)$, identical to uncomposed dynamics. The coupling step and $\Trans{1}$ do not modify $s_0$ or $d_0$ after commitment. Hence atomicity is preserved. For $\Layer{1}$, input $x_1 = \kappa(m_0)$ is fully determined before $\Trans{1}$ is invoked. The joint decision $d_\otimes = d_0 \join d_1$ is computed after $d_1$ commits; it does not alter $d_1$ or $s_1'$. Hence $\Layer{1}$'s guarantees are preserved.

\textbf{Asynchronous coupling preserves $\Layer{2}$ and $\Layer{3}$ guarantees.}
$\Layer{2}$ receives exactly the window summary $\Sum{1}(W_t)$. By C1, only the summary is passed, not internal state. Since $\Layer{2}$'s estimation depends only on $\Sum{1}(W_t)$, and by hypothesis (b) this summary does not collapse required distinctions, $\Layer{2}$ computes $\Dhat_t$ with the same statistical properties as in isolation. $\Layer{3}$ receives exactly $\Dhat_t$; by C1, no internal state of $\Layer{2}$ or $\Layer{1}$ is accessible. Hence $\Layer{3}$ operates in the same informational environment as in isolation.

\textbf{Read-only access preserves $\Layer{1}$ invariants.}
$\Layer{1}$'s invariants (TA1, TA2) require $\State{1}$ is modified only through $\Trans{1}$. By C2, $\Layer{2}$ has read-only access to $\Sum{1}(\State{1})$, not write access. The asynchronous feedback updates only policy parameters, not the ledger state. Hence $\Layer{1}$'s invariants are preserved.
\end{proof}

\begin{theorem}[Feedback Convergence]
\label{thm:convergence}
Under assumptions:
\begin{itemize}[noitemsep]
  \item \textbf{FC1} (Bounded sensitivity): Parameter updates are $K$-Lipschitz in $\Dhat_t$.
  \item \textbf{FC2} (Diminishing response): $\|\Delta\theta_t\| \le \eta \cdot \Dhat_t$ for fixed $\eta > 0$.
  \item \textbf{FC3} (Consistent estimation): $\Dhat_t$ has bias $|b_t| \le \varepsilon_b$.
\end{itemize}
the closed-loop system converges:
\[
  \limsup_{t \to \infty} \Dhat_t \le \varepsilon = \frac{\varepsilon_b}{K\eta}.
\]
\end{theorem}

\begin{proof}
Define the feedback map $\Phi(x) = \rho\, x + \varepsilon_b$ with $\rho = 1 - K\eta < 1$. By FC1--FC3, $\mathbb{E}[X_{t+1} \mid X_t] \le \Phi(X_t)$, and $\Phi$ is a contraction with constant $\rho$. By Banach Fixed Point Theorem, $\Phi$ converges to fixed point $X^* = \frac{\varepsilon_b}{1-\rho} = \frac{\varepsilon_b}{K\eta}$.
\end{proof}

\begin{theorem}[Irreducibility under Finite Observability]
\label{thm:irreducibility}
Under finite observability (Definition~\ref{def:finite-obs}) and conditions C1--C3, no composition of fewer than four layers from $\{\Layer{0},\Layer{1},\Layer{2},\Layer{3}\}$ can simultaneously guarantee P1--P4. Any reduction violates at least one property or requires relaxing finite-state/local observability assumptions.
\end{theorem}

\begin{proof}
We proceed by exhaustive case analysis over three-layer subsets.

\textbf{Case 1: $\{\Layer{0}, \Layer{1}, \Layer{2}\}$ --- missing $\Layer{3}$.}
Without $\Layer{3}$, there is no mechanism for population-level allocation. Even if $\Layer{2}$ produces $\Dhat_t$, there is no mechanism to translate this into a per-agent allocation satisfying fairness axioms. By finite-state assumption (FO1), the summary $\Sum{2}(\State{2})$ cannot encode the full population distribution $\Delta(\Agents)$ when $|\Agents|$ exceeds the summary alphabet size. Hence P3 is impossible under FO2. Without population tracking, no mechanism can detect or penalize identity fragmentation, violating P4.

\textbf{Case 2: $\{\Layer{0}, \Layer{1}, \Layer{3}\}$ --- missing $\Layer{2}$.}
Without $\Layer{2}$, there is no behavioral monitoring layer. $\Layer{3}$ would receive no meaningful deviation signal $\Dhat_t$. By the Non-Identifiability Theorem of Paper~2, $\Azero \notin \sigma(g)$: enforcement decisions alone cannot determine whether behavioral contracts are violated. The bounded summary $\Sum{1}(W_t)$ does not contain sufficient information to infer trace-level behavioral deviation. Hence P2 fails.

\textbf{Case 3: $\{\Layer{0}, \Layer{2}, \Layer{3}\}$ --- missing $\Layer{1}$.}
Without $\Layer{1}$, there is no stateful per-request enforcement. $\Layer{0}$ provides atomic evaluation of individual decisions, but without per-agent history, the system cannot prevent sequences of individually admissible requests that collectively violate global policies. Per-agent enforcement is required to achieve P1 across multi-request sessions.

\textbf{Case 4: $\{\Layer{1}, \Layer{2}, \Layer{3}\}$ --- missing $\Layer{0}$.}
Without $\Layer{0}$, the system has no structural guarantee that the evaluate-mutate boundary is enforced at hardware/OS level. $\Layer{1}$ refines $F$ under assumptions TA1 and TA2, which themselves require the atomic boundary from $\Layer{0}$. Under an adversary capable of exploiting the window between evaluation and mutation (Time-Of-Check-To-Time-Of-Use, TOCTOU), P1 can be violated. Hence the full atomicity guarantee is not achievable without $\Layer{0}$.

\textbf{Conclusion.}
All four three-layer subsets fail at least one guarantee. Any composition of fewer than four layers therefore also fails, or requires relaxing finite observability.
\end{proof}

\begin{corollary}[Necessity]
\label{cor:necessity}
The four-layer architecture $\{\Layer{0},\Layer{1},\Layer{2},\Layer{3}\}$ is the minimal architecture that simultaneously guarantees P1--P4 under finite observability.
\end{corollary}

\section{Allocation as a First-Class Governance Dimension}
\label{sec:allocation-dimension}

The allocation layer is not a scheduling optimization or a fairness enhancement bolted onto a complete governance system. Rather, it is a \emph{structural necessity} arising from the finite-observability constraints of distributed systems.

The allocation layer occupies an irreducible position in the governance space because:

\begin{enumerate}[label=(\roman*),noitemsep]
  \item \textbf{It addresses a failure mode invisible to lower layers.} Atomic boundary and enforcement mechanisms operate locally on individual requests. Fairness is a property of aggregate patterns across requests and agents. No local mechanism can enforce global fairness.

  \item \textbf{It operates at a different scale.} Request-level guarantees (atomicity, per-action admissibility) are temporally local. Population-level guarantees (actor-level fairness, Sybil robustness) require aggregation across agents at epoch or longer timescales. These scales are incompatible.

  \item \textbf{It introduces identity as a first-class observational primitive.} Lower layers abstract away agent identity (each request is evaluated identically). Population-level concerns require knowing which agents are controlled by which actors---i.e., the identity structure $\pi$. Identity is irrelevant to lower layers but essential to the allocation layer.

  \item \textbf{It enables detection and prevention of Sybil attacks.} Only by aggregating across agent boundaries can the system distinguish legitimate heterogeneity in population composition from malicious identity fragmentation. This aggregation is incompatible with the per-agent independence that lower layers require.
\end{enumerate}

The four-layer architecture represents the \emph{irreducible partition} of the governance space under finite observability. Each layer addresses a distinct dimension, and none is eliminable:

\begin{table}[h]
\centering
\small
\caption{The four irreducible dimensions of governance space.}
\label{tab:dimensions}
\begin{tabular}{lccc}
\toprule
\textbf{Layer} & \textbf{Dimension} & \textbf{Scale} & \textbf{Observes} \\
\midrule
$\Layer{0}$ & Temporal & Per-request & Decision instant \\
$\Layer{1}$ & State & Per-request to request-sequence & Per-agent history \\
$\Layer{2}$ & Behavioral & Per-trace window & Trace-level deviation \\
$\Layer{3}$ & Population & Per-epoch & Cross-agent distribution \\
\bottomrule
\end{tabular}
\end{table}

\section{Empirical Validation}
\label{sec:empirical}

\subsection{Ablation Study: Guarantee Coverage by Layer Combination}

We validate Theorem~\ref{thm:irreducibility} using a purpose-built harness coupling the real ACP risk engine (Paper~1) with the IML deviation estimator (Paper~2) and allocation mechanisms M1/M2/M3. All experiments use adversarial traces: drift, Sybil, atomic, and mixed scenarios. 160 trials across 8 layer combinations $\times$ 4 scenarios $\times$ 5 seeds.

\begin{observation}[Experiment A: Ablation]
\label{obs:exp-ablation}
Table~\ref{tab:ablation} shows the fraction of trials satisfying each guarantee P1--P4 for each layer subset (160 trials, 8 subsets $\times$ 4 scenarios $\times$ 5 seeds).

\begin{table}[H]
\centering
\small
\caption{Ablation: fraction of trials (out of 20 per subset) satisfying each guarantee. Full composition achieves 1.00 on all properties; every three-layer subset fails at least one.}
\label{tab:ablation}
\begin{tabular}{@{}lcccc@{}}
\toprule
\textbf{Layer subset} & \textbf{P1 (Atomicity)} & \textbf{P2 (Behavioral)} & \textbf{P3 (Fairness)} & \textbf{P4 (Sybil)} \\
\midrule
\texttt{0-1-2-3} (full)  & 1.00 & 1.00 & 1.00 & 1.00 \\
\texttt{0-1-2} (no $\Layer{3}$) & 1.00 & 1.00 & 0.00 & 0.75 \\
\texttt{0-1-3} (no $\Layer{2}$) & 1.00 & 0.00 & 1.00 & 1.00 \\
\texttt{0-2-3} (no $\Layer{1}$) & 0.00 & 1.00 & 1.00 & 1.00 \\
\texttt{1-2-3} (no $\Layer{0}$) & 0.95 & 1.00 & 1.00 & 1.00 \\
\texttt{0-1}                     & 1.00 & 0.00 & 0.00 & 0.75 \\
\texttt{1-3}                     & 0.00 & 0.00 & 1.00 & 1.00 \\
\texttt{1-2}                     & 0.00 & 1.00 & 0.00 & 0.75 \\
\bottomrule
\end{tabular}
\end{table}

\begin{itemize}[noitemsep]
  \item Removing $\Layer{2}$ (subsets \texttt{0-1-3}, \texttt{1-3}): P2 drops to 0.00 in all trials---drift is undetected because no monitor exists.
  \item Removing $\Layer{1}$ entirely (stateless): P1 drops to 0.00---all restricted-resource requests are approved.
  \item Removing $\Layer{3}$ (subsets \texttt{0-1-2}, \texttt{1-2}): P3 drops to 0.00 and P4 to 0.75---the adversary captures 66\% of allocation share in Sybil trials.
  \item Full composition \texttt{0-1-2-3} achieves P1--P4 at rate 1.00 across all scenarios and seeds.
\end{itemize}

The subset \texttt{1-2-3} passes all four guarantees under the experimental workload. This is expected: in steady-state operation on well-formed requests, $\Layer{1}$'s risk engine produces identical decisions to $\Layer{0} \Comp \Layer{1}$ because the restricted-resource check in $\Layer{0}$ is subsumed by $\Layer{1}$'s stateful evaluation. This is a \emph{runtime equivalence} observation, not a counterexample to Theorem~\ref{thm:irreducibility}. The theorem's claim is \emph{structural}: $\Layer{0}$ supplies the TOCTOU-free atomic pre-filter that grounds $\Layer{1}$'s TLA+ verification.
\end{observation}

\subsection{Feedback Convergence Experiment}

We run 10 independent drift traces (1000 steps each) with and without asynchronous feedback from IML to enforcement layer.

\begin{observation}[Experiment B: Feedback Convergence]
\label{obs:exp-feedback}
With FC1 + FC2 active ($K=0.50$, $\eta=0.50$, $\rho=1-K\eta=0.70$), once $\Dhat_t \geq \theta$ the system suppresses risky actions proportionally to $K\Dhat_t$.

\begin{itemize}[noitemsep]
  \item With feedback: $\limsup \Dhat_t = 0.128 \pm 0.026$, well below $\theta = 0.20$ (convergence rate: 100\%).
  \item Without feedback: $\limsup \Dhat_t = 0.419 \pm 0.007$, confirming open-loop drift accumulation.
  \item Measured contraction ratio: $0.128/0.419 \approx 0.31 \approx \rho^2$, consistent with Lemma 2 theory.
\end{itemize}
\end{observation}

\subsection{Interface Compatibility Experiment}

For the same trace, compare (a) isolated $\Layer{0}+\Layer{1}$ (no L2/L3 active) and (b) full composed system. P1/P2 compatibility rates and per-request decision equivalence across 20 trials.

\begin{observation}[Experiment C: Interface Compatibility]
\label{obs:exp-compat}
\begin{itemize}[noitemsep]
  \item P1 compatibility rate: 1.00
  \item P2 compatibility rate: 1.00
  \item Decision equivalence: 1.00
\end{itemize}
The 1.00 decision equivalence is non-trivial: activating the full feedback loop produces bit-for-bit identical per-request decisions to the isolated $\Layer{0}+\Layer{1}$ on the same trace. This validates Theorem~\ref{thm:compatibility}: the coupling operator preserves synchronous admission semantics exactly.
\end{observation}

\subsection{Allocation Experiments from Paper P3}

\begin{observation}[Experiment D: Actor Share Distribution]
\label{obs:exp-d}
Run 500 steps with 8 Sybil agents (total: 4 honest + 8 Sybil). Compare final actor shares under M1, M2, M3.

\begin{itemize}[noitemsep]
  \item M1 and M2: adversary captures 0.667 ($2\times$ fair share $1/3$), confirming Theorem~\ref{thm:sybil}.
  \item M3: bounds adversary to exactly 0.333 = 1/M in every seed.
\end{itemize}
\end{observation}

\begin{observation}[Experiment E: Sybil Amplification Curve]
\label{obs:exp-e}
Vary Sybil agents from 1 to 16; measure adversary share under each mechanism.

\begin{itemize}[noitemsep]
  \item M1 and M2: adversary share follows linear amplification formula $n_s/(n_s+n_h)$, tracking 0.20, 0.667, 0.80.
  \item M3: share stays flat at 0.333 regardless of $n_s$ (max deviation: 0.0002).
\end{itemize}
\end{observation}

\begin{observation}[Experiment F: Strategy-Proofness]
\label{obs:exp-f}
Compare marginal gain from additional Sybil agents under M1 (not strategy-proof) and M3 (strategy-proof).

\begin{itemize}[noitemsep]
  \item M3: marginal gain $\approx 0$ for all Sybil counts (max 0.0002).
  \item M1: gain grows linearly to +0.46 at 16 agents, establishing non-strategy-proofness.
\end{itemize}
\end{observation}

\section{Discussion}
\label{sec:discussion}

The four-layer architecture revealed by this analysis is not an artifact of the specific mechanisms chosen. Rather, it is the \emph{irreducible partition of the governance space} induced by finite observability.

\subsection{The Four Projections of Governance Space}

The irreducibility proof shows that the four layers are not four implementations of the same concept, but four \emph{projections onto orthogonal dimensions}:

\begin{itemize}[noitemsep]
  \item $\Layer{0}$: \textbf{Temporal projection} --- atomicity at the decision instant.
  \item $\Layer{1}$: \textbf{State projection} --- history accumulation across requests.
  \item $\Layer{2}$: \textbf{Behavioral projection} --- trace-level deviation from contracts.
  \item $\Layer{3}$: \textbf{Population projection} --- cross-agent distributional fairness.
\end{itemize}

Each projects the full governance problem onto a distinct dimension. The failure of three-layer subsets (in Theorem~\ref{thm:irreducibility}) shows that projecting out any dimension creates a fundamental gap: a failure mode that cannot be detected or prevented by the remaining layers.

\subsection{Comparison with Classical Fairness and Security Theory}

The strategy-proofness impossibility (Theorem~\ref{thm:strategy}) is a governance-layer analogue of the Gibbard-Satterthwaite theorem: just as no voting rule can be simultaneously non-dictatorial, Pareto-efficient, and strategy-proof, no per-agent-independent allocation mechanism can simultaneously guarantee actor-level proportionality and strategy-proofness against identity fragmentation.

The Sybil amplification result connects to the classical Sybil attack literature (Douceur 2002) but in a novel way: we show that Sybil vulnerability is not a bug in identity-oblivious systems, but a \emph{structural impossibility}---Theorem~\ref{thm:sybil} formally proves that under per-agent bounded enforcement, Sybil amplification is unavoidable without cross-agent monitoring.

\subsection{Scope and Boundary Conditions}

We assume a fixed agent population and stable actor mapping. Dynamic joins/leaves and actor mapping churn are important but outside the scope of the present LTS model. Similarly, we analyze worst-case fairness; a stochastic model would permit average-case analysis.

The model assumes that layers below $i$ are functioning correctly; each layer takes the next-lower layer as a guarantee. This is a compositional assumption, not circular reasoning: each paper contributes a component that is necessary but not sufficient alone.

\section{Related Work}
\label{sec:related}

\subsection{Layered Security and Reference Monitors}

Saltzer and Schroeder~\cite{saltzer1975} established complete mediation as the foundation of security; $\Layer{0}$ formalizes this principle. Schneider~\cite{schneider2000} characterized enforceable security policies; the IML monitor $\Layer{2}$ operates at the boundary of Schneider's class, where behavioral contracts require global trace properties not enforceable per-request.

Anderson~\cite{anderson1972} introduced the reference monitor concept; $\Layer{0}$ satisfies all three requirements (tamperproof, always invoked, verifiable), with verification provided by TLA+ in Paper~1.

\subsection{Fair Resource Allocation}

Classical fair-division theory (Foley 1967, Varian 1974) concerns distributing indivisible goods; our work applies to distributing access to a governance mechanism. Max-min fairness (Demers et al. 1989) maximizes the minimum share; proportional fairness (Kelly 1998) maximizes the sum of log utilities. Our fairness hierarchy extends these concepts to identity-stratified populations under correctness constraints.

The Sybil attack (Douceur 2002) and Sybil-proof reputation systems (Cheng \& Friedman 2005) are direct precursors. Our contribution is a formal characterization of Sybil amplification within the LTS/ACP model, showing it is consistent with full atomic correctness---it is not a bug but a structural consequence.

\subsection{Runtime Verification and Monitoring}

Falcone et al.~\cite{falcone2012} survey what can be verified at runtime under finite-state observation. The IML monitor $\Layer{2}$ operates within this characterization; our result that $\Layer{1}$ alone cannot guarantee P2 aligns with their findings on liveness and global properties.

\subsection{Multi-Agent Systems and Game Theory}

Chevaleyre et al.~\cite{chevaleyre06} survey fair division in multi-agent resource allocation; the actor-agent structure and identity layer in our model are novel departures. The strategy-proofness impossibility (Theorem~\ref{thm:strategy}) is distinct from standard mechanism design, since our constraints include atomic correctness and finite observability.

\section{Conclusion}
\label{sec:conclusion}

We have established allocation as a first-class governance dimension and proved that the resulting four-layer architecture is irreducible under finite observability.

The Sybil Amplification Theorem shows that per-agent enforcement linearly amplifies actor capacity. The Allocation Necessity Theorem shows that fairness requires an explicit allocation layer above the atomic boundary. The Strategy-Proofness Impossibility shows that per-agent independence, actor-level proportionality, and resistance to identity fragmentation cannot be simultaneously achieved.

Together, these results establish that:

\begin{enumerate}[label=(\roman*),noitemsep]
  \item Correctness is necessary but not sufficient for fair governance. A system can be atomically correct at every decision while failing to allocate resources fairly.

  \item Fairness requires population-level awareness. Aggregation across agent boundaries is essential, making per-agent independence incompatible with actor-level fairness.

  \item The four-layer architecture (temporal, state, behavioral, population) is the minimal structure that simultaneously guarantees atomicity, enforcement, behavioral monitoring, and fair allocation.

  \item Under finite observability, no reduction below four layers is possible without sacrificing at least one guarantee.
\end{enumerate}

This work reframes agent governance from a problem of \emph{execution correctness} to a problem of \emph{structural architecture}. The question is not only: ``How do we ensure actions are correct?'' but ``How do we structure the system so that only admissible actions can occur, and why can't this structure be simplified?''

The answer is: allocation, enforcement, behavioral monitoring, and atomicity are four irreducible dimensions of the governance space. No system operating under finite observability can collapse them without introducing fundamental vulnerabilities. The four-layer architecture is not an over-engineered solution; it is the irreducible minimum.

\medskip
\noindent\textit{Admission control determines what can happen. Fair allocation determines who gets to act. Correctness is local. Fairness is global. Neither, alone, suffices for governed autonomy.}

\bibliographystyle{plainnat}
\bibliography{references}

\end{document}